\documentclass[sigconf]{acmart}
\usepackage{booktabs}
\usepackage{array}
\usepackage{multirow}
\usepackage{makecell}
\usepackage{subcaption}
\usepackage{amsmath}
\usepackage{xurl}
\setcopyright{none}
\acmConference[SAC '27]{The 42nd ACM/SIGAPP Symposium on Applied Computing}{April 5--9, 2027}{Gwangju, South Korea}
\acmYear{2027}
\copyrightyear{2027}
\acmDOI{}
\acmISBN{}
\begin{document}
\title[PseuDRAGON: Knowledge-Guided Local LLMs]{PseuDRAGON: Knowledge-Guided Local LLMs for Financial-Data Pseudonymization in Closed Networks}
\author{Gwangui Hong}
\authornote{These authors contributed equally to this work.}
\affiliation{\institution{KAIST}\city{Daejeon}\country{South Korea}}
\affiliation{\institution{KFTC}\city{Seoul}\country{South Korea}}
\email{recordame@kaist.ac.kr}

\author{Seonho An}
\authornotemark[1]
\affiliation{\institution{KAIST}\city{Daejeon}\country{South Korea}}
\email{asho1@kaist.ac.kr}

\author{Seonkyu Lim}
\authornote{Corresponding authors.}
\affiliation{\institution{KFTC}\city{Seoul}\country{South Korea}}
\email{sklim@kftc.or.kr}

\author{KyungTae Lim}
\authornotemark[2]
\affiliation{\institution{KAIST}\city{Daejeon}\country{South Korea}}
\email{ktlim@kaist.ac.kr}

\author{Min-Soo Kim}
\authornotemark[2]
\affiliation{\institution{KAIST}\city{Daejeon}\country{South Korea}}
\email{minsoo.k@kaist.ac.kr}

\renewcommand{\shortauthors}{Hong et al.}
\begin{abstract}
Staff pseudonymizing financial data in closed networks must interpret abbreviated column names, identify personal information, and choose processing methods. Local language models can assist, but may give different labels for the same input. PseuDRAGON keeps confirmed judgments and corrections outside a fixed model and reuses them through exact-name lookup and learned rules. For expert review, the workflow searches institutional policy, national privacy law, then international sources. Across 92 columns from one institution, rules increase reuse under the same approval checks from 75 to 83 labels without reuse errors when all training judgments are available. With 25\% of the training judgments, withholding non-PII rules for unseen names eliminated the extra PII misses in replay at the cost of lower reuse. Revising or withdrawing approvals rebuilds derived rules, and tests on synthetic records verify that selected methods produce the specified outputs. In three paired runs, the history-first baseline cut automated identification time by 83.0\% on average versus no reuse.
\end{abstract}
\maketitle
\section{Introduction}
\label{sec:introduction}
Sharing financial data for analysis requires protecting personal information. At the institution we study, staff pseudonymize ledger data inside a network isolated from the Internet. Korea's SaaS network-separation exemption excludes unique identification information and personal credit information~\citep{fsc2026saas}. Financial authorities elsewhere address risk-based network segmentation~\citep{nfra2024datasecurity,mas2021trm,ffiec2021authentication,sama2017cybersecurity,ec2024dora}. Staff must interpret columns, identify personal information, and choose a method.

\begin{figure*}[t]
\centering
\includegraphics[width=.86\textwidth,trim=0 42 0 35,clip]{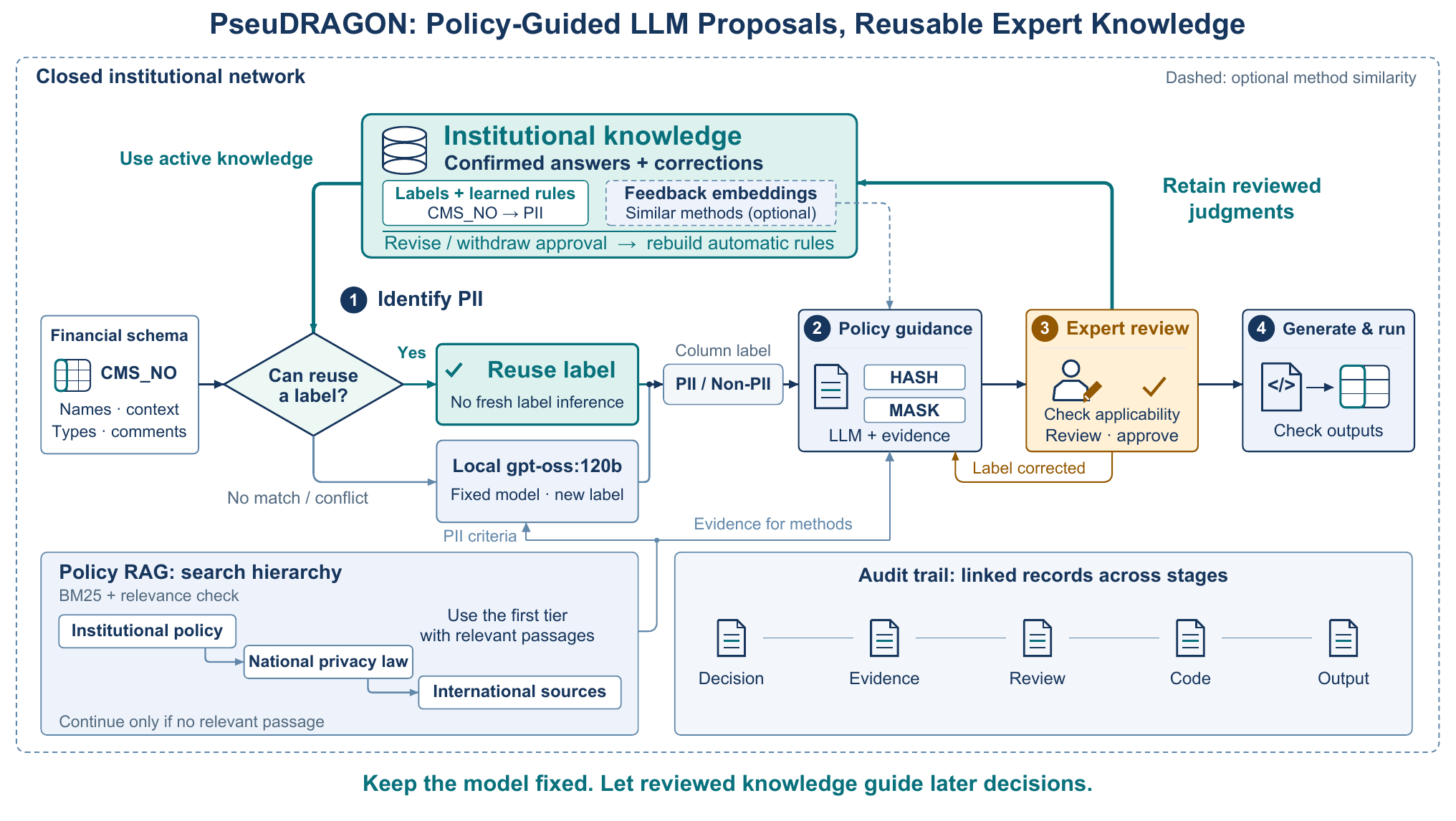}
\Description{A closed-network workflow with PII identification, policy guidance, expert review, program generation, and execution checks. Matching columns reuse saved labels and learned rules; unmatched or conflicting columns receive a fresh label from the fixed local gpt-oss:120b. Both routes lead to method review. A solid loop saves confirmed judgments and corrections for later requests. Source-ordered RAG searches institutional policy, national privacy law, then international sources. It uses the first tier with relevant passages to supply evidence for labeling and method proposals. An optional dashed branch uses feedback embeddings to suggest similar methods. Audit records link decisions, evidence, review, code, and outputs.}
\caption{Closed-network RegTech workflow: source-ordered policy RAG and knowledge reuse with a fixed model. Audit records link column decisions, policy evidence, expert review, selected code, and checked outputs.}
\label{fig:pseudragon-pipeline}
\end{figure*}

Internal abbreviations can obscure what a column means~\citep{kang2003opaque,zhang2023nameguess}. The cash management service number \texttt{CMS\_NO} in Figure~\ref{fig:pseudragon-pipeline} looked like an operational code until staff confirmed that it links to a customer one-to-one. Hashing can preserve links for analysis~\citep{enisa2019pseudonymisation}, and masking can meet institutional display rules.

LLMs support tasks in finance~\citep{li2023llmfinance}. Repeated model calls can return different outputs for the same input~\citep{atil2025nondeterminism,ouyang2025nondeterminism}. When a column recurs, fresh inference can reopen a judgment staff have already checked. Context can aid identification~\citep{ntwali2025structured}, RAG provides evidence~\citep{lewis2020rag}, and caches and feedback memories reuse earlier outputs~\citep{gptcache2023,madaan2022memprompt,dalvi2022teachme}. Reuse also depends on whether the supporting approval remains active.

Review yields reusable knowledge even when a label requires no correction. In a replay, keeping corrections alone reuses 4 labels, while keeping all reviewed decisions reuses 75. PseuDRAGON\footnote{Source code: \url{https://github.com/recordame/PseuDRAGON}.} is a closed-network regulatory technology (RegTech) workflow~\citep{fsb2020regtech} that stores those decisions outside a fixed local model, checks active approvals for conflicts, and rebuilds rules when approvals change. The audit record links evidence, selected code, and outputs.

We evaluate 92 columns from nine reconstructed financial tables with synthetic records. Held-out replays measure where rules extend reuse and where incomplete knowledge causes misses. Approval injection and withdrawal test how errors propagate and how rebuilding removes their effects. We time identification and check outputs against independent reference operations.

\begin{samepage}
Our contributions follow the workflow:
\begin{itemize}
\item A closed-network workflow that links policy evidence, expert review, and checked programs (Section~\ref{sec:industry-method}).
\item Reuse of confirmed judgments and corrections, with rule rebuilding when approvals change (Section~\ref{sec:review-knowledge-reuse}).
\item An evaluation of rule transfer, approval errors and withdrawal, and execution of selected methods (Section~\ref{sec:industry-evaluation}).
\end{itemize}
\end{samepage}

\section{Related Work}
\label{sec:related}
\paragraph{PII and column semantics.} Presidio offers entity recognizers, patterns, and adjustable anonymization operations~\citep{presidio2023}. For structured data, Ntwali et al. combine a field's name and values with other field names and dataset descriptions~\citep{ntwali2025structured}. Telkamp and Hulsebos distinguish context within a dataset from external domain information~\citep{telkamp2025contextual}. Schema matching also uses relationships between attributes to find matches when names are unclear~\citep{kang2003opaque}. Sherlock labels columns from their values~\citep{hulsebos2019sherlock}. Sato and Doduo also use table context~\citep{zhang2020sato,suhara2022doduo}. LLMs can annotate column types with few or no examples~\citep{korini2023cta,feuer2024archetype}, and NameGuess expands abbreviated column names~\citep{zhang2023nameguess}. In our setting, interpreting a column leads to a reviewed decision that can be reused when the column recurs.

\paragraph{Anonymization.} Language-model anonymization studies how to rewrite protected content. SEAL learns to anonymize and critique text, then improves it through repeated revisions~\citep{kim2025self}. Hou et al. separate sensitive-entity detection, replacement-candidate generation, and replacement, then restore private entities in the returned text~\citep{hou2025general}. PseuDRAGON instead reviews operations on structured columns. Pseudonymization techniques differ in linkability, utility, and recovery~\citep{enisa2019pseudonymisation}, and released records can still be re-identified by linking them to other data~\citep{sweeney2002kanonymity}. We check whether the selected method, parameters, and code produce the specified output.

\paragraph{Retrieval-augmented generation.} RAG uses retrieved material to guide answers~\citep{lewis2020rag,izacard2021fid}. Legal question answering can retrieve statutory articles and cite supporting paragraphs~\citep{louis2023legalrag}. RAG-HAT uses hallucination detection and preference tuning to improve faithfulness~\citep{song2024raghat}. Reliability-Aware RAG estimates source reliability to prioritize retrieval and combine evidence~\citep{hwang2025reliabilityrag}. Our retrieval instead searches policy sources in a fixed order with relevance checks.

\paragraph{Feedback and knowledge reuse.} Human--AI interaction guidelines emphasize explanations, correction, and visible effects of feedback~\citep{amershi2019guidelines}. Wu et al. connect interaction design and evaluation with learning from human feedback~\citep{wu2023hitl}. Semantic caches also reuse earlier answers: GPTCache returns stored model responses for matching queries, avoiding new generation~\citep{gptcache2023}. Feedback memories keep user corrections outside a frozen model and retrieve them as context for similar queries~\citep{madaan2022memprompt,dalvi2022teachme}. Weak supervision combines heuristic labeling functions of unknown accuracy~\citep{ratner2017snorkel}. We keep confirmed decisions as well as corrections outside a fixed model, derive rules only from reviewed approvals, and test how conflicts and approval changes affect reuse.

\section{Closed-Network RegTech Workflow}
\label{sec:industry-method}
Figure~\ref{fig:pseudragon-pipeline} shows the four stages: PII identification (Section~\ref{sec:policy-retrieval}), method proposal and expert review with knowledge reuse (Section~\ref{sec:review-knowledge-reuse}), and program generation (Section~\ref{sec:program-generation}). An optional branch uses feedback embeddings~\citep{reimers2019sbert} to suggest similar methods.

\subsection{Problem setup}
\label{sec:problem-setup}
An operator selects tables and states the intended use. Each column $c$ has a name $\nu(c)$, a data type, and a comment. The workflow assigns each column a label $y(c)\in\{\text{PII},\text{non-PII}\}$ and a method with parameters, such as hashing, encryption, or masking for PII columns, and keeping or deleting non-PII columns. A reviewer checks column meanings, requirements, and proposed operations in table context, then confirms or corrects each proposal. Each reviewed label becomes an approval $a$ with name $\nu(a)$, label $y(a)$, explanation, and evidence. Let $\mathcal{A}$ be the active approvals and $\mathcal{R}$ the automatic rules learned from them (Section~\ref{sec:review-knowledge-reuse}), together with any manual rules. With the fixed local model $M$ and retrieved evidence $E(c)$, PseuDRAGON labels a column as
\begin{equation}
\hat{y}(c)=
\begin{cases}
M\bigl(c,E(c)\bigr) & \text{if same-name approvals in }\mathcal{A}\text{ disagree},\\
y(a) & \text{else if some }a\in\mathcal{A}\text{ has }\nu(a)=\nu(c),\\
y(r) & \text{else if }r\text{ is the first rule in }\mathcal{R}\text{ matching }\nu(c),\\
M\bigl(c,E(c)\bigr) & \text{otherwise}.
\end{cases}
\label{eq:reuse}
\end{equation}
The second and third cases depend only on $c$, $\mathcal{A}$, and $\mathcal{R}$, so covered columns receive deterministic labels when inputs and stores are unchanged. Revising or withdrawing an approval updates $\mathcal{A}$ and rebuilds the automatic rules in $\mathcal{R}$.

\begin{table*}[t]
\centering
\caption{The 25 column names in the 92 columns, with schema meanings, reference PII labels, and the number of tables containing each name (32 PII and 60 non-PII columns). Labels are constant within each name.}
\label{tab:full-schema}
\footnotesize
\setlength{\tabcolsep}{3pt}
\begin{minipage}[t]{.49\textwidth}
\vspace{0pt}
\centering
\begin{tabular}{@{}lp{4.2cm}cr@{}}
\toprule
Column name & Meaning & PII & Table count\\
\midrule
\texttt{CUST\_ID} & Customer identifier & Yes & 2\\
\texttt{WD\_ACNO} & Withdrawal account number & Yes & 7\\
\texttt{DPSTR\_NM} & Depositor name & Yes & 6\\
\texttt{DPS\_ACNO} & Deposit account number & Yes & 4\\
\texttt{CMS\_NO} & Cash management service no. & Yes & 5\\
\texttt{USER\_SEQ\_NO} & User sequence number & Yes & 3\\
\texttt{CI} & Identity-linking information & Yes & 3\\
\texttt{FIN\_INTT\_SVC\_NO} & Financial transfer service no. & Yes & 2\\
\texttt{STD\_DT} & Standard date & No & 9\\
\texttt{WD\_BANK\_CD} & Withdrawal bank code & No & 8\\
\texttt{WD\_MN} & Withdrawal amount & No & 8\\
\texttt{TRNS\_DTIME} & Transaction date/time & No & 5\\
\texttt{CRNO} & Corporate registration no. & No & 5\\
\bottomrule
\end{tabular}
\end{minipage}
\hfill
\begin{minipage}[t]{.49\textwidth}
\vspace{0pt}
\centering
\begin{tabular}{@{}lp{4.2cm}cr@{}}
\toprule
Column name & Meaning & PII & Table count\\
\midrule
\texttt{RCMS\_BSWR\_CL\_CD} & Recipient inquiry category & No & 4\\
\texttt{DPS\_BANK\_CD} & Deposit bank code & No & 4\\
\texttt{WD\_DT} & Withdrawal date & No & 4\\
\texttt{TRNS\_MED\_CL\_CD} & Transaction medium & No & 2\\
\texttt{TRNS\_CL\_CD} & Transaction category & No & 2\\
\texttt{FEE} & Fee & No & 2\\
\texttt{DEALCO\_CD} & Handling institution code & No & 2\\
\texttt{WD\_RSLT\_CD} & Withdrawal result code & No & 1\\
\texttt{INQR\_DTIME} & Inquiry date/time & No & 1\\
\texttt{INQR\_CL\_CD} & Inquiry category & No & 1\\
\texttt{BLNC} & Balance & No & 1\\
\texttt{REQ\_DT} & Request date & No & 1\\
\bottomrule
\end{tabular}
\end{minipage}
\end{table*}

\subsection{Interpreting columns and requirements}
\label{sec:policy-retrieval}
Stage 1 labels each column with Eq.~\eqref{eq:reuse}. For columns routed to the model, $M$ receives retrieved evidence, table and column names, comments, and data types, and returns a PII or non-PII label.

Search follows institutional policy, national privacy law, then international sources. Column descriptions retrieve passages in each category, and the model checks their relevance to the column and intended use. Let $P_k(c)$ be the relevant passages in category $k\in\{1,2,3\}$. Search stops at the first relevant category, giving $E(c)=P_{k^*}(c)$ for $k^*=\min\{k:P_k(c)\neq\emptyset\}$. If none supplies a relevant passage, the system records the gap and proceeds without retrieved evidence. Later categories may contain requirements that this search omits. The order governs retrieval, and reviewers must check applicable requirements across sources before approval.

BM25 ranks each category's passages and adds nearby text as specified in Table~\ref{tab:experiment-settings}. Relevance checks use task and passage text, without retrieval scores or evaluation labels. Rejected passages do not carry forward; invalid responses cannot stop search. Records keep category and passage IDs; method parameters require review.

\subsection{Review and knowledge reuse}
\label{sec:review-knowledge-reuse}
Stage 2 records methods, parameters, reasons, sources, and code. For account-display requests, a bounded parser reads explicit source instructions. The evaluated instruction permits hiding at least three digits at the end or before the final digit, producing two alternatives. Each records the position, digit count, mask character, missing-value handling, and source, with a matching template. Unrecognized or conflicting instructions produce no compiled candidate. Experts review these alternatives alongside model candidates.

\begin{figure*}[t]
\centering
\begin{subfigure}[b]{.34\textwidth}
\centering
\includegraphics[width=\linewidth,trim=0 0 241.89 13,clip]{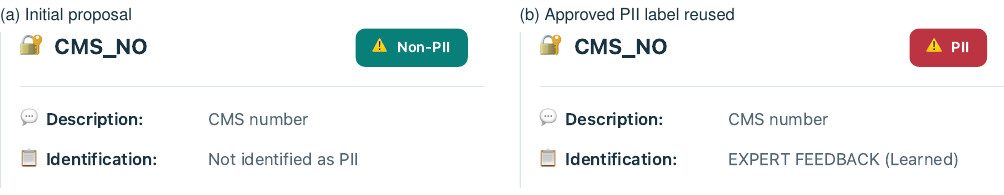}
\caption{Initial proposal}
\end{subfigure}\hspace{1.5em}
\begin{subfigure}[b]{.34\textwidth}
\centering
\includegraphics[width=\linewidth,trim=241 0 0 13,clip]{figures/review-reuse.pdf}
\caption{Approved PII label reused}
\end{subfigure}\\[4pt]
\begin{subfigure}[b]{.29\textwidth}
\centering
\includegraphics[width=\linewidth,trim=0 0 481.23 18,clip]{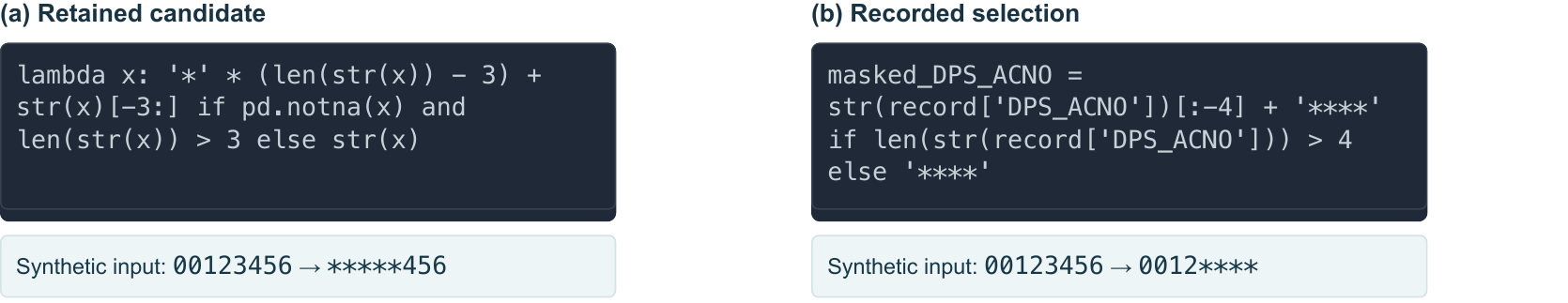}
\caption{Retained candidate}
\end{subfigure}\hspace{1.5em}
\begin{subfigure}[b]{.29\textwidth}
\centering
\includegraphics[width=\linewidth,trim=410 0 69.23 18,clip]{figures/mask-code-aligned.pdf}
\caption{Recorded selection}
\end{subfigure}
\Description{Top: Two cropped PDF views of CMS_NO from the running web interface, preserving its displayed text and styles. The first view shows the non-PII badge from the initial proposal. The second shows the PII badge and identifies expert feedback as the source of the reused label after demonstration approval. The example uses reconstructed schema metadata and recorded proposals, rather than fresh model inference. Bottom: Two vector captures of the current web source-code editor populated with recorded MASK choices for HF_TRNS_TRAN.DPS_ACNO. The candidate code keeps the last three digits, whereas the recorded selection masks the last four characters. Output annotations show synthetic input 00123456 becoming *****456 and 0012****. These are recorded choices replayed for illustration.}
\caption{Recorded examples: (a, b) \texttt{CMS\_NO} labels before and after a demonstration approval whose explanation records the customer mapping; (c, d) \texttt{DPS\_ACNO} MASK code and annotated outputs for the same synthetic input.}
\label{fig:review-reuse}
\label{fig:mask-code}
\end{figure*}

Stage 3 lets reviewers correct labels and adjust methods; changed labels regenerate the policy before approval. Feedback stores labels and explanations; rules store patterns and support. Figure~\ref{fig:review-reuse}(a,b) shows the initial non-PII label and the approved PII label on reuse.

Rules learn from confirmed judgments and corrections. Each name supplies a prefix through its first underscore and suffixes from each underscore. An initial candidate must have at least two supporting entries and meet the scoring threshold:
\begin{equation}
s=\min\Bigl(1,\frac{n}{N_y}+0.1u\Bigr)\geq 0.8,
\label{eq:rule-score}
\end{equation}
where $n$ counts supporting entries, $N_y$ counts entries with label $y$, and $u$ distinct supporting names. This heuristic combines label coverage and name diversity; it is not a correctness probability.

New rules are checked against all matching training entries, including rare counterexamples. Opposite labels in more than 0.10 of these entries remove the rule. Uncovered names get exact patterns. Refinement merges same-label exact patterns with a common suffix and at least two supporting names, bypassing the initial score threshold. For example, reviewed non-PII judgments for \texttt{STD\_DT} and \texttt{WD\_DT} can merge into a \texttt{\_DT} rule for \texttt{REQ\_DT}. Longer suffixes use parent examples and the same support and score thresholds. Equal example sets keep the longer suffix.

A merged suffix with $m$ exact-name examples scores $\min(1,0.7+0.1m)$; a longer refinement with $n$ matches among $N_r$ parent examples scores $\min(1,n/N_r+0.1n)$. Rules apply by descending priority: $\lfloor 100s\rfloor$ for $s$ rounded to three decimals, plus 10 for longer refinements. Exact-name rules have priority 50; ties retain stored order.

Name and pattern reuse applies within an active scope of reviewed column meanings and method requirements. Revised or withdrawn approvals trigger rebuilding automatic rules from active feedback, preserving manual rules. As in retraining-based machine unlearning~\citep{bourtoule2021unlearning}, rebuilding removes a withdrawn approval's influence. When meanings or requirements change, the operator archives old stores and starts new ones, carrying forward only reviewed judgments and manual rules. Review also saves accepted methods and edited code as preferences for later proposals.

\subsection{Generating pseudonymization programs}
\label{sec:program-generation}
Stage 4 combines selected code or templates into one program per table. Running it transforms an in-memory copy of the synthetic records and exports the output without changing the source.

Code must match its method and parameters. Supported masking and generalization edits update both; other edits require an explicit, unverified declaration. Both three-digit compiled masks preserve separators and missing inputs, hiding all digits when there are at most three. On \texttt{00123456}, they yield \texttt{00123***} or \texttt{0012***6}. Selection rechecks source identity, parameters, and program structure. All selected operations preserve missing inputs.

For \texttt{DPS\_ACNO} in the home and firm banking table \texttt{HF\_TRNS\_TRAN}, the initial and reference labels are PII. In Figure~\ref{fig:mask-code}(c,d), the retained original MASK shows the last three digits, while the selected MASK hides the last four characters. On \texttt{00123456}, the candidate exposes both protected ranges; the selection meets the display rule. Section~\ref{sec:industry-execution} tests both. Records link proposals and selections to methods, parameters, reasons, evidence, and supporting knowledge. Verification fingerprints the stores, selected programs, and their outputs.

\section{Evaluation}
\label{sec:industry-evaluation}
We measure the accuracy and consistency of fresh PII labels. We test how incorrect or withdrawn approvals affect reuse and check program outputs against independent reference operations.

\subsection{Experimental setup}
\label{sec:setup}
\paragraph{Data and reference.} Table~\ref{tab:table-inventory} lists the nine reconstructed financial tables and their purposes. Of their 92 columns, 32 are PII and 60 non-PII in the reference. Table~\ref{tab:full-schema} summarizes reference PII labels for columns that identify an individual customer or account holder: names, customer and linkage identifiers, account numbers, and customer-linked service numbers. Corporate identifiers and transaction details such as dates and balances are non-PII in this task; thus \texttt{DPS\_ACNO} is positive and \texttt{CRNO} negative. A study author built the reference with responsible business staff, checking column meanings, underlying data, and one-to-one links to individuals. It was finalized before blind practitioner evaluation and inference. Tests use synthetic records preserving these column meanings. Labels guide identification; the institution approves release.

\begin{table}[ht]
\centering
\caption{Financial tables used in the evaluation.}
\label{tab:table-inventory}
\footnotesize
\setlength{\tabcolsep}{3pt}
\begin{tabular}{@{}lp{3.8cm}r@{}}
\toprule
Table name & Purpose & Columns\\
\midrule
\texttt{CD\_TRNS\_TRAN} & CD/ATM withdrawals and transfers & 13\\
\texttt{CMS\_REQ\_TRAN} & CMS withdrawal requests & 7\\
\texttt{CMS\_RES\_TRAN} & CMS withdrawal results & 8\\
\texttt{GR\_JC\_TRAN} & Direct-debit collection requests & 10\\
\texttt{GR\_NJ\_TRAN} & Payer direct debits & 10\\
\texttt{HF\_TRNS\_TRAN} & Home and firm banking transfers & 14\\
\texttt{OB\_INQR\_TRAN} & Open banking inquiries & 7\\
\texttt{OB\_TRNS\_TRAN} & Open banking transfers & 11\\
\texttt{PI\_WD\_LEDG} & Debit-transfer ledger & 12\\
\midrule
Total & & 92\\
\bottomrule
\end{tabular}
\end{table}

\begin{table*}[t]
\centering
\caption{Experiment settings and controls.}
\footnotesize
\label{tab:experiment-settings}
\begin{tabular}{@{}m{2.7cm}m{14.6cm}@{}}
\toprule
Component & Configuration (document aliases I1--G1)\\
\midrule
Institutional sources & I1: display rules (revision marker 2022-04-20; 7 chunks); I2: privacy rules (revised 2025-10-15; 223 chunks).\\\addlinespace[1pt]
National sources & N1: Personal Information Protection Act No.~20897~\citep{pipa2025} (effective 2025-10-02; 417 chunks); N2: privacy-policy drafting guidance (April 2025; 537 chunks).\\\addlinespace[1pt]
International source & G1: GDPR~\citep{gdpr2016} translation (EU 2016/679; 2,192 chunks); translation edition date unrecorded.\\\addlinespace[1pt]
Passage retrieval & Five documents, 3,376 uniquely identified chunks; character-bigram/trigram BM25 ($k_1=1.2$, $b=0.75$). Each category supplies up to three positive-score passages with adjacent same-document text; overlapping groups are merged.\\\addlinespace[1pt]
Model generation & \texttt{gpt-oss:120b}; temperature 0; low reasoning effort; output limits 4,096 tokens for proposals and 2,048 for relevance checks. Relevance caching stays within a repetition or timing condition.\\\addlinespace[1pt]
Held-out replay & Withheld families: 27, 15, 20, 18, and 12 columns. Stores use other families' reference-confirmed labels; static rules and embeddings are disabled. Methods share fixed fallbacks: initial-run predictions or each source-ordered set.\\\addlinespace[1pt]
Source-ordered runs & Three runs without stored judgments or heuristics: 92 columns and 11 account-display requests with two names. Inputs: names, types, and comments; no cell values; fixed display requirement. Total: 208 relevance checks, 309 proposal calls.\\\addlinespace[1pt]
Timing runs & History-first baseline versus no reuse; three paired runs, three workers per condition; order no reuse--reuse, reuse--no reuse, no reuse--reuse. Each condition starts with an empty relevance cache; stores stay fixed.\\
\bottomrule
\end{tabular}
\end{table*}

\paragraph{Practitioners.} Eight practitioners judged columns without seeing reference labels, system outputs, or peer judgments. Their responses were kept separate from feedback. Six were current or former staff at KFTC (Korea Financial Telecommunications \& Clearings Institute, Korea), and two were from FSI (Financial Security Institute, Korea).

\paragraph{Model.} The main evaluation runs use \texttt{gpt-oss:120b}~\citep{openai2025gptoss120b}, served on an NVIDIA H100 (80 GB) inside the institutional network. Table~\ref{tab:experiment-settings} lists the sources, retrieval, and generation settings. Three kinds of runs have separate records: the \emph{initial run}, the first institutional response, which also sees one synthetic row per table; three metadata-only \emph{source-ordered runs}; and three paired \emph{timing runs}.

\paragraph{Reuse routes.} The 92 columns share 25 names, making name caching a useful baseline. \emph{No reuse} labels every column with the model. The \emph{name cache} reuses exact-name approvals only. The \emph{history-first baseline} consults exact-name history before rules, without the approval check. \emph{Cache + rules} is PseuDRAGON's default route (Eq.~\eqref{eq:reuse}). Unless stated otherwise, the name cache and cache + rules share the same approval checks, and uncovered or conflicting columns take labels from three fixed prediction sets: the saved source-ordered runs of Section~\ref{sec:latest-proposals}.

\paragraph{Held-out protocol.} We withhold each schema family and build from the others' reference labels, without new practitioner approvals. Families share table-name prefixes or header-set Jaccard similarity of at least 0.8. Grouping uses no labels; each column is tested once. Held-out labels affect neither stores nor predictions.

\paragraph{Metrics.} We report precision (P), recall (R), and F1 against the reference. Reused labels come from stored knowledge through exact names or rules; fresh labels come from the model, and their counts exclude subsequent method proposals. Reuse errors are reused labels that disagree with the reference, separate from final prediction errors. Cases count injections that introduce errors; added errors count the held-out labels made wrong by those injections.

\subsection{Identification and consistency}
\label{sec:latest-proposals}

\begin{table}[ht]
\centering
\caption{Column identification against the reference.}
\label{tab:financial-identification}
\small
\begin{tabular}{@{}llrrr@{}}
\toprule
Decision source & Group & P & R & F1 \\
\midrule
E1 & \multirow{3}{*}{KFTC (current)} & 1.000 & 0.781 & 0.877 \\
E2 & & 1.000 & 0.625 & 0.769 \\
E3 & & 1.000 & 0.781 & 0.877 \\
\cmidrule(lr){1-5}
E4 & \multirow{3}{*}{KFTC (former)} & 1.000 & 0.594 & 0.745 \\
E5 & & 1.000 & 0.750 & 0.857 \\
E6 & & 1.000 & 1.000 & 1.000 \\
\cmidrule(lr){1-5}
E7 & \multirow{2}{*}{FSI} & 0.604 & 0.906 & 0.725 \\
E8 & & 0.865 & 1.000 & 0.928 \\
\midrule
\multicolumn{2}{@{}l}{Strict practitioner majority} & 1.000 & 0.781 & 0.877 \\
\multicolumn{2}{@{}l}{Initial run} & 1.000 & 0.906 & 0.951 \\
\multicolumn{2}{@{}l}{Reviewed state, same items} & 1.000 & 1.000 & 1.000 \\
\bottomrule
\end{tabular}
\end{table}

Majority voting can hide minority judgments in subjective labeling~\citep{van2024acal}. A strict majority requires at least five of eight positive judgments. In Table~\ref{tab:financial-identification}, the majority identifies 25 of 32 positive columns without false positives against the reference; the initial run identifies 29. Scored against majority judgments, the same predictions give F1 0.926. Two practitioners with ten years of experience differ in the categories they protect. All 5 false positives from one practitioner are corporate-registration columns excluded by the institutional reference. E6 matches all reference labels but neither set the reference nor supplied feedback.

The same-item check tests retention after review of the initial run. After reference disclosure, all rerun labels match: 77 columns use stored knowledge (70 rules and 7 feedback records), and the model supplies the remaining 15 labels.

Three source-ordered runs label the same 92 columns without stored judgments or heuristics. F1 is 1.000, 1.000, and 0.985. Labels agree for 91/92 columns; the third run mislabels one balance column as positive. The reported evidence changes for 18 columns under the same inference settings and retrieval procedure. Nine switch between source categories, and nine switch between a source category and no retrieved evidence. The labels from these runs supply the fixed reuse fallbacks.

\begin{samepage}
An earlier paired check compares the weighted retriever with the source-ordered pipeline on the same 92 columns, with one prediction per column in each arm. Both arms use the same model, schema metadata, and generation settings. Both arms correctly label all 32 PII and 60 non-PII columns. The change also affects passage ranking, context windows, and relevance checks, so the comparison evaluates the two retrieval pipelines as a whole. Checking whether early stopping misses applicable requirements calls for measuring coverage across sources.
\par
\end{samepage}

\subsection{Reusing reviewed decisions}
\label{sec:retaining-decisions}

We compare the name cache and cache + rules under the same approval checks and held-out protocol in Table~\ref{tab:industry-heldout}.

\begin{table}[ht]
\centering
\caption{Reuse with all training judgments and identical approval checks across three fixed prediction sets.}
\label{tab:industry-heldout}
\small
\begin{tabular}{@{}lrrrr@{}}
\toprule
Method & Exact & Rules & \makecell{Fresh\\labels} & \makecell{Reuse\\errors}\\
\midrule
\multicolumn{5}{@{}l}{\textit{Schema family withheld}}\\
\quad No reuse & 0 & 0 & 92 & 0\\
\quad Name cache & 75 & 0 & 17 & 0\\
\quad \textbf{Cache + rules} & \textbf{75} & \textbf{8} & \textbf{9} & \textbf{0}\\
\midrule
\multicolumn{5}{@{}l}{\textit{Family withheld; same-name records also removed}}\\
\quad Name cache & 0 & 0 & 92 & 0\\
\quad \textbf{Cache + rules} & \textbf{0} & \textbf{39} & \textbf{53} & \textbf{0}\\
\bottomrule
\end{tabular}
\end{table}

With families withheld, Table~\ref{tab:industry-heldout} shows reuse rising from 75 to 83 columns without reuse errors after adding rules. Name coverage grows from 14 to 19 of 25. The extra 8 are among 17 columns with names absent from training: 2 positives and 6 negatives across 5 names. F1 stays 1.000, 1.000, and 0.985 across the fixed prediction sets; the only error comes from a model fallback.

In a stricter replay within the same institution, we also remove all same-name training records before rebuilding each query's store. The rules correctly label 39 of 92 columns across 12 names (8 PII, 31 non-PII). Exact-name reuse assumes that the name has the same meaning for identification across tables. In these schemas each name has a consistent reference label.

\begin{samepage}
We also rebuild stores as training judgments accumulate in 20 fixed orders (Table~\ref{tab:knowledge-growth}). At 25\% of the judgments, a non-PII rule for \texttt{WD\_} mislabels two held-out \texttt{WD\_ACNO} columns in two orders. With no account-number counterexample in the store, the rule overrides correct fallback predictions. A separate replay defers non-PII rules for names absent from the store. At 25\%, mean additional false negatives fall from 0.20 to 0 and mean reuse falls from 62.20 to 51.35 columns. The same policy reduces reuse from 83 to 77 columns when each training store contains all its judgments.
\par
\end{samepage}

Replaying initial-run predictions tests what confirmed answers add to corrections. Here the history-first baseline holds either corrections only or all reviewed selections. Corrections alone reuse 4 labels without fixing any of the 3 misses; all selections reuse 75 and fix all 3. The name cache fixes 2; rules also reach an unseen service-number name (Section~\ref{sec:knowledge-maintenance}). Across the three fixed prediction sets, corrections alone cover no columns.

\begin{figure}[ht]
\centering
\begin{subfigure}[b]{.9\columnwidth}
\centering
\includegraphics[width=\linewidth,trim=0 0 249.06 15,clip]{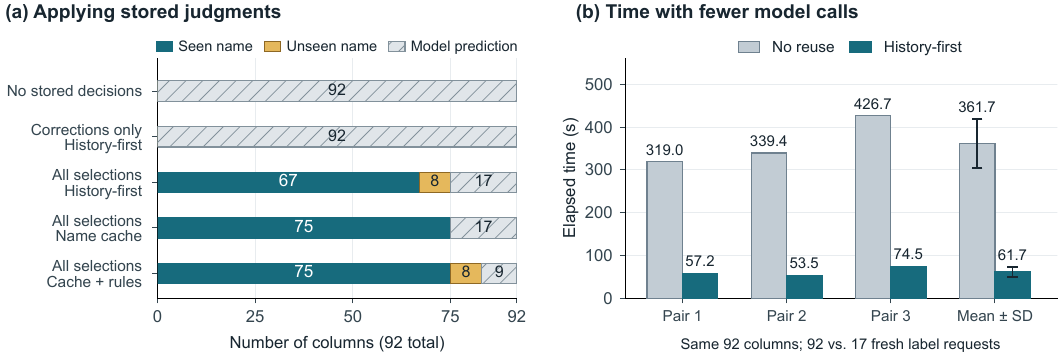}
\caption{Applying stored judgments}
\end{subfigure}\\[4pt]
\begin{subfigure}[b]{.9\columnwidth}
\centering
\includegraphics[width=\linewidth,trim=261 0 0 15,clip]{figures/reuse-efficiency-en.pdf}
\caption{Automated identification time}
\end{subfigure}
\Description{Two panels show reuse by names present in training and measured identification time. Across fixed-prediction replays, corrections alone cover zero columns. The history-first baseline covers 67 seen-name and eight unseen-name columns; the name cache covers 75 seen-name columns; their combination covers 75 seen-name and eight unseen-name columns. A separate live comparison measures three pairs with 92 label-generation calls without reuse and 17 with the recorded history-first baseline. Mean times are 361.7 and 61.7 seconds; error bars on the means show sample standard deviations.}
\caption{Reuse with a fixed model. (a) Fixed-prediction replay of the history-first baseline and name-cache variants. (b) Timing runs of the history-first baseline versus no reuse; mean error bars show sample SD ($n=3$).}
\label{fig:reuse-efficiency}
\end{figure}

Timing runs compare no reuse with the history-first baseline, using training-family stores without the added approval check. In Figure~\ref{fig:reuse-efficiency}(b), label calls fall from 92 to 17 and mean times from $361.7\pm57.2$ to $61.7\pm11.2$ seconds (sample SD), a mean paired reduction of 83.0\%. Timing covers lookup, retrieval, generation, parsing, and audit writes. Initialization and store building, human review, and method selection are excluded.

Timing-run F1 is 1.000, 0.985, and 1.000 without reuse, and 1.000 with reuse. The only difference is \texttt{OB\_INQR\_TRAN.BLNC} in run 2, which both conditions predict afresh with the same model, inputs, and settings: the same request yields different labels.

\subsection{Ablation: approvals and reuse}
\label{sec:knowledge-maintenance}
Table~\ref{tab:knowledge-ablation} evaluates the approval checks before history lookup and during learning. Each of 92 columns appears in four training stores: 368 injection positions. Flipping one label at each position across three fixed prediction sets yields 1,104 cases.

Approval checks reduce added errors in both rule routes to 4 errors in 2 cases per set, without changing clean coverage. Each failure has one wrong approval and no stored conflict to reveal it. The name cache, the history-first baseline with the approval check, and cache + rules agree on held-out labels, including fallbacks, across stages and prediction sets. Clean reuse is 75, 75, and 83, respectively. In all 1,104 cases, rebuilding after withdrawal gives each method the same routes as a new store built from remaining approvals. Restoring the correct approval recovers clean routes.

\begin{table}[ht]
\centering
\caption{Approval checks and rule rebuilding: added-error cases and errors per 368 injection positions. Counts match across three fixed prediction sets.}
\label{tab:knowledge-ablation}
\small
\begin{tabular}{@{}lrr@{}}
\toprule
Condition & Cases & Errors\\
\midrule
\multicolumn{3}{@{}l}{\textit{History-first}}\\
\quad Before approval check & 208 & 322\\
\quad With approval check & 2 & 4\\
\midrule
\multicolumn{3}{@{}l}{\textit{Cache + rules}}\\
\quad Conflict deferral removed & 48 & 75\\
\quad \textbf{Conflict deferral enabled} & \textbf{2} & \textbf{4}\\
\midrule
\multicolumn{3}{@{}l}{\textit{Name cache}}\\
\quad Approval check enabled & 2 & 4\\
\midrule
\multicolumn{3}{@{}l}{\textit{After withdrawal (rule-based routes)}}\\
\quad Approval removed; rules retained & 2 & 4\\
\quad \textbf{Approval removed; rules rebuilt} & \textbf{0} & \textbf{0}\\
\bottomrule
\end{tabular}
\end{table}

Five \texttt{CMS\_NO} approvals and one \texttt{USER\_SEQ\_NO} approval support a PII \texttt{\_NO} rule for unseen \texttt{FIN\_INTT\_SVC\_NO}. Changing one \texttt{CMS\_NO} approval to non-PII removes the rule after rebuilding; withdrawing it restores the rule. The target's PII label stays unchanged while its route moves from reuse to fixed fallback and back.

New counterexamples can remove rules and reduce reuse. Table~\ref{tab:knowledge-growth} tracks this change. Reuse is error-free at higher training fractions.

\begin{table}[ht]
\centering
\caption{Mean reuse counts for 92 columns across 20 judgment orders and three fixed prediction sets.}
\label{tab:knowledge-growth}
\small
\begin{tabular}{@{}rrrr@{}}
\toprule
Judgments & History-first & Name cache & \textbf{Cache + rules}\\
\midrule
0\% & 0 & 0 & 0\\
25\% & 61.40 & 50.45 & \textbf{62.20}\\
50\% & 76.25 & 67.00 & \textbf{77.85}\\
75\% & 79.15 & 72.50 & \textbf{82.15}\\
100\% & 75.00 & 75.00 & \textbf{83.00}\\
\bottomrule
\end{tabular}
\end{table}

The rule settings affect reuse and the risk of leaving missed PII untreated. Table~\ref{tab:reuse-sensitivity} varies minimum support (2 or 3), score threshold (0.7 or 0.9), and conflict limit (0 or 0.2) over the same 20 orders. At 25\%, all reuse errors are missed PII. With all training judgments available, each setting correctly reuses 83 labels.

\begin{table}[ht]
\centering
\caption{Cache + rules sensitivity. Partial-store entries give ranges of 20-order means; full-store entries give counts.}
\label{tab:reuse-sensitivity}
\small
\begin{tabular}{@{}lrrr@{}}
\toprule
\multirow{2}{*}{Setting} & \multicolumn{2}{c}{25\% judgments} & 100\%\\
\cmidrule(lr){2-3}
 & Reused & Wrong & Reused\\
\midrule
Default & 62.20 & 0.20 & 83\\
Support 2; four settings & 61.70--62.25 & 0.20--0.30 & 83\\
Support 3; four settings & 58.30--58.90 & 0.20--0.30 & 83\\
\bottomrule
\end{tabular}
\end{table}

\begin{samepage}
\subsection{Executing selected methods}
\label{sec:industry-execution}
Across three runs, 33 account-display requests each yield a raw MASK candidate and two source-compiled alternatives. Of 33 raw candidates, 32 pass the three digit-only strings. Table~\ref{tab:industry-integrated} compares unchanged raw code with compiled alternatives on the same 17 inputs. Raw code fails on short strings, separators, and missing values. For example, \texttt{12-34-56} becomes \texttt{12-34***}, hiding only two digits. Small test sets can overstate the correctness of generated code~\citep{liu2023evalplus}. Raw candidates also lack parameter declarations, preventing selection. The 66 compiled alternatives have matching parameters and code and pass all 1,122 comparisons using automatic test selections.
\par
\end{samepage}

\begin{table}[ht]
\centering
\caption{MASK checks across three runs. Passing candidates meet every assigned input check; contract checks match methods, parameters, and code.}
\label{tab:industry-integrated}
\small
\begin{tabular}{@{}lrrr@{}}
\toprule
Candidate & Inputs & Passing & \makecell{Contract\\pass}\\
\midrule
Raw MASK & 3 & 32/33 & 0/33\\
Raw MASK & 17 & 0/33 & 0/33\\
\textbf{Compiled} & \textbf{17} & \textbf{66/66} & \textbf{66/66}\\
\bottomrule
\end{tabular}
\end{table}

We separately check the retained candidates and selections from Section~\ref{sec:program-generation} on six digit-only inputs. All 4 selected masks meet the display rule in 24/24 program--input cases. Of the candidates, 9 runnable ones pass 0/54 cases and 2 others have syntax errors.

We then run nine table programs three times each on read-only synthetic data. The 92 selections use hashing for 25 columns, masking for 4, interval generalization for 3, and KEEP for 60. With 1,000 rows per table, all 27 outputs match separately implemented reference operations over 92,000 unique cells (276,000 comparisons). Reference functions follow operation definitions, independently of the normalizer and selected source code.

All 92 columns pass method/parameter and missing-value checks. All 1,288 boundary-cell and 276 expanded missing-value comparisons pass, and inputs remain unchanged. Boundary inputs include short or empty strings, leading zeros, multilingual text, emoji, control characters, and interval boundaries.

A synthetic check tests whether the selected customer-ID hashes preserve links across two tables with 84 transactions from 12 fictitious customers. The outputs preserve all 144 cross-table pairs and each customer's transaction count and total amount. Changing one key preserves the match count while changing the linked pairs. These links support analysis, and linking released records with outside data can still allow re-identification~\citep{sweeney2002kanonymity}.

\section{Limitations and Ethical Considerations}
\label{sec:limitations}
\paragraph{Limitations.} We evaluate one institution's reconstructed schemas and synthetic records. Staff-checked business data support the reference, without external adjudication or a dated original label copy. Same-item replay tests retention; family replay tests reference-based rule transfer with fixed predictions. Execution checks selected operations on tested inputs. Broader code correctness, re-identification resistance, reviewer productivity, and deployment effects remain outside this evaluation.

\paragraph{Ethics.}\label{sec:industry-ethics} The eight practitioners consented to participate. No information that could identify them was collected or recorded. No separate institutional ethics review was obtained. Institutional execution uses synthetic records and read-only source access. Execution checks do not certify legal compliance or privacy protection.

\section{Conclusion}
\label{sec:conclusion}
PseuDRAGON stores confirmed labels and corrections outside the local model. Exact lookup and learned rules return the same labels for covered columns as long as the inputs and stored knowledge stay the same. If staff revise or withdraw an approval, the system rebuilds the learned rules. When a reviewed column recurs, the next analyst can reuse its approved label and inspect the evidence.

\bibliographystyle{ACM-Reference-Format}
\bibliography{pseudragon}
\end{document}